%% file: neurips_2025.tex
\documentclass{article}

\usepackage[utf8]{inputenc} 
\usepackage[T1]{fontenc}    

\input{preamble/00_begin}
\input{preamble/01_neurips}
\input{preamble/02_text}
\input{preamble/03_figures}
\input{preamble/04_mathematics}

\input{preamble/05_notation}
\input{preamble/98_draft}

\input{preamble/99_end}

\title{Limits of PRM-Guided Tree Search for\\Mathematical Reasoning with LLMs}

\author{%
  Tristan Cinquin\thanks{Work done while interning at the Vector Institute.} \\
  University of Tübingen\\
  \texttt{tristan.cinquin@uni-tuebingen.de} \\
  \And
  Geoff Pleiss \\
  University of British Columbia \& Vector Institute \\
  \texttt{geoff.pleiss@stat.ubc.ca} \\
  \And
  Agustinus Kristiadi \\
  Western University \& Vector Institute \\
  \texttt{akristi@uwo.ca} \\
}

\begin{document}

\maketitle

\begin{abstract}
\input{00_abstract}
\end{abstract}

\input{01_intro}
\input{02_background}
\input{03_methods}
\input{04_results}
\input{05_related_work}
\input{06_conclusion}

\begin{ack}
\input{99_ack}
\end{ack}

\bibliography{neurips_2025}

\newpage
\input{07_appendix}

\end{document}

%% file: preamble/00_begin.tex
\usepackage{ifdraft}

%% file: preamble/01_neurips.tex
\PassOptionsToPackage{numbers, compress}{natbib}

\usepackage[dblblindworkshop,final]{neurips_2025}

\workshoptitle{MATH-AI}



%% file: preamble/02_text.tex
\usepackage{microtype}

\usepackage[inline]{enumitem}

\usepackage[draft=false]{hyperref}
\usepackage[numbered]{bookmark}  

\usepackage[numbers,compress]{natbib}
\usepackage{doi}

\usepackage[nottoc, section]{tocbibind}

\usepackage[dvipsnames]{xcolor}

\hypersetup{
colorlinks,
linkcolor=NavyBlue
,citecolor=PineGreen
,filecolor=Mulberry
,urlcolor=NavyBlue
,menucolor=BrickRed
,runcolor=Mulberry
}

%% file: preamble/03_figures.tex
\usepackage{subfig}

\usepackage[final]{graphicx}

\graphicspath{%
  {figures/}%
}

\usepackage{tikz}

\usetikzlibrary{%
  calc,%
  positioning,%
  arrows.meta,%
}

\usepackage{booktabs}
\usepackage{multirow}
\usepackage{wrapfig}

\usepackage{algorithm}
\usepackage[indLines=true]{algpseudocodex}

%% file: preamble/04_mathematics.tex
\usepackage{amsmath}
\usepackage{mathtools}

\usepackage{amssymb}
\usepackage{bm}
\usepackage{nicefrac}
\usepackage{dsfont}

\usepackage{amsthm}
\usepackage{thmtools}

\usepackage{etoolbox}
\usepackage{xparse}

\allowdisplaybreaks    

\numberwithin{equation}{section}  

\newcommand*{\defeq}{\coloneqq}  

\DeclarePairedDelimiterX{\set}[1]\{\}{%

#1}

\newcommand*{\setsym}[1]{\ensuremath{{\mathbb{#1}}}}

\newcommand*{\sA}{\setsym{A}}
\newcommand*{\sB}{\setsym{B}}

\newcommand*{\sF}{\setsym{F}}

\newcommand*{\sS}{\setsym{S}}
\newcommand*{\sT}{\setsym{T}}

\DeclarePairedDelimiter{\card}{\lvert}{\rvert}

\newcommand*{\closure}[1]{\operatorname{cl} \left( #1 \right)}

\renewcommand*{\vec}[1]{{\bm{#1}}}

\newcommand*{\vtheta}{\vec{\theta}}

\newcommand*{\vphi}{\vec{\phi}}

\newcommand*{\vpsi}{\vec{\psi}}

\NewDocumentCommand{\range}{s m}{%
  \operatorname{ran}
  \IfBooleanTF{#1}{\left(}{(}
  #2
  \IfBooleanTF{#1}{\right)}{)}
}

\NewDocumentCommand{\kernel}{s m}{%
  \operatorname{ker}
  \IfBooleanTF{#1}{\left(}{(}
  #2
  \IfBooleanTF{#1}{\right)}{)}
}

\newcommand*{\mat}[1]{{\bm{#1}}}

\makeatletter
\DeclarePairedDelimiterXPP{\@norm}[2]{}{\lVert}{\rVert}{\ifstrempty{#2}{}{_{#2}}}{#1}
\NewDocumentCommand{\norm}{s O{} m O{}}{%
  \IfBooleanTF{#1}{%
    \@norm*{#3}{#4}%
  }{%
    \@norm[#2]{#3}{#4}%
  }%
}
\makeatother

\newcommand*{\dualop}{'}

\makeatletter
\DeclarePairedDelimiterXPP{\@inprod}[3]{}{\langle}{\rangle}{\ifstrempty{#3}{}{_{#3}}}{#1, #2}
\NewDocumentCommand{\inprod}{s O{} m m O{}}{%
  \IfBooleanTF{#1}{%
    \@inprod*{#3}{#4}{#5}%
  }{%
    \@inprod[#2]{#3}{#4}{#5}%
  }%
}
\makeatother

\NewDocumentCommand{\cfns}{m o}{%
  C (#1\IfValueT{#2}{, #2})
}

\NewDocumentCommand{\cdfns}{O{0} m o}{%
  C^{#1} (#2\IfValueT{#3}{, #3})
}

\NewDocumentCommand{\holdersp}{m o m}{%
  C^{#1\IfValueT{#2}{, #2}}(\closure{#3})
}

\RenewDocumentCommand{\L}{m o}{%
  \ensuremath{
    L_{#1}
    \IfNoValueF{#2}{\left( #2 \right)}
  }
}

\NewDocumentCommand{\sobolev}{o m o}{%
  \IfNoValueTF{#1}{
    H^{#2}
  }{
    W^{#1,#2}
  }
  \IfNoValueF{#3}{\left( #3 \right)}
}

\NewDocumentCommand{\sobolevtest}{o m o}{%
  \sobolev[#1]{#2}_0
  \IfNoValueF{#3}{\left( #3 \right)}
}

\newcommand*{\linop}[1]{\mathcal{#1}}

\newcommand*{\linfctls}[1]{{\bm{\linop{#1}}}}

\makeatletter
\DeclarePairedDelimiter{\@evallinop@oparg}{[}{]}
\DeclarePairedDelimiter{\@evallinop@fnarg}{(}{)}
\NewDocumentCommand{\evallinop}{m s O{} m s O{} d()}{%
  #1%
  \IfBooleanTF{#2}{%
    \@evallinop@oparg*{#4}%
  }{%
    \@evallinop@oparg[#3]{#4}%
  }%
  \IfValueT{#7}{%
    \IfBooleanTF{#5}{%
      \@evallinop@fnarg*{#7}%
    }{%
      \@evallinop@fnarg[#6]{#7}%
    }%
  }%
}
\makeatother

\NewDocumentCommand{\linopat}{m s O{} m d()}{%
  \IfBooleanTF{#2}{%
    \evallinop{\linop{#1}}*{#4}(#5)%
  }{%
    \evallinop{\linop{#1}}[#3]{#4}(#5)%
  }%
}

\NewDocumentCommand{\linfctlsat}{m s O{} m d()}{%
  \IfBooleanTF{#2}{%
    \evallinop{\linfctls{#1}}*{#4}(#5)%
  }{%
    \evallinop{\linfctls{#1}}[#3]{#4}(#5)%
  }%
}

\newcommand*{\diff}[2][1]{%
  \mathrm{d} #2\ifstrequal{#1}{1}{}{^{#1}}%
}

\newcommand*{\pdiff}[2][1]{%
  \partial #2\ifstrequal{#1}{1}{}{^{#1}}%
}

\NewDocumentCommand{\deriv}{s O{1} m m}{%
  \frac{%
    \mathrm{d}\ifstrequal{#2}{1}{}{^{#2}}\IfBooleanT{#1}{#4}
  }{%
    \diff[#2]{#3}
  }
  \IfBooleanF{#1}{#4}
}

\NewDocumentCommand{\derivat}{s O{1} m m m}{%
  \left.
    \IfBooleanTF{#1}{%
      \deriv*[#2]{#3}{#4}
    }{%
      \deriv[#2]{#3}{#4}
    }
  \right|_{#3 = #5}
}

\NewDocumentCommand{\dderiv}{m m}{%
  \partial_{#1} #2
}

\NewDocumentCommand{\dderivat}{m m o m}{%
  \IfNoValueTF{#3}{%
    \dderiv{#1}{#2} \left( #4 \right)
  }{%
    \left.
    \dderiv{#1}{#2}
    \right_{#3 = #4}
  }
}

\NewDocumentCommand{\pderiv}{s O{1} m m}{%
  \frac{%
    \partial\ifstrequal{#2}{1}{}{^{#2}}\IfBooleanT{#1}{#4}
  }{%
    \pdiff[#2]{#3}
  }
  \IfBooleanF{#1}{#4}
}

\NewDocumentCommand{\pderivat}{s O{1} m m o m}{%
  \left.
    \IfBooleanTF{#1}{%
      \pderiv*[#2]{#3}{#4}
    }{%
      \pderiv[#2]{#3}{#4}
    }
  \right|_{\IfNoValueTF{#5}{#3}{#5} = #6}
}

\NewDocumentCommand{\mpderiv}{s O{1} m m}{%
  \frac{%
    \partial\ifstrequal{#2}{1}{}{^{#2}}\IfBooleanT{#1}{#4}
  }{%
    #3
  }
  \IfBooleanF{#1}{#4}
}

\NewDocumentCommand{\mpderivat}{s O{1} m m m m}{%
  \left.
    \IfBooleanTF{#1}{%
      \mpderiv*[#2]{#3}{#4}
    }{%
      \mpderiv[#2]{#3}{#4}
    }
  \right|_{#5 = #6}
}

\NewDocumentCommand{\mipderiv}{s m o m o}{%
  \IfNoValueTF{#3}{
    \mathrm{D}^{#2}
    \IfBooleanTF{#1}{\left[ #4 \right]}{#4}
    \IfNoValueF{#5}{\left( #5 \right)}
  }{
    \IfNoValueF{#5}{\left.}
    \frac{%
      \partial^{\lvert #2 \rvert}\IfBooleanT{#1}{#4}
    }{%
      \pdiff[#2]{#3}
    }
    \IfBooleanF{#1}{#4}
    \IfNoValueF{#5}{\right\vert_{#3 = #5}}
  }
}

\NewDocumentCommand{\jacobian}{o m o}{%
  \IfNoValueTF{#1}{%
    \mat{\mathrm{D}} #2 \IfNoValueF{#3}{\left( #3 \right)}
  }{%
    \IfNoValueTF{#3}{
      \mat{\mathrm{D}}_{#1} #2
    }{
      \left. \mat{\mathrm{D}}_{#1} #2 \right|_{#1\IfNoValueF{#3}{= #3}}
    }
  }
}

\NewDocumentCommand{\gradient}{o m o}{%
  \IfNoValueTF{#1}{%
    \nabla #2 \IfNoValueF{#3}{\left( #3 \right)}
  }{%
    \IfNoValueTF{#3}{
      \nabla_{#1} #2
    }{
      \left. \nabla_{#1} #2 \right|_{#1\IfNoValueF{#3}{= #3}}
    }
  }
}

\NewDocumentCommand{\divergence}{m o}{%
  \operatorname{div} \left( #1 \right) \IfNoValueF{#2}{\left( #2 \right)}
}

\NewDocumentCommand{\hessian}{o m o}{%
  \IfNoValueTF{#1}{%
    \mat{\mathrm{H}} #2 \IfNoValueF{#3}{\left( #3 \right)}
  }{%
    \IfNoValueTF{#3}{
      \mat{\mathrm{H}}_{#1} #2
    }{
      \left. \mat{\mathrm{H}}_{#1} #2 \right|_{#1\IfNoValueF{#3}{= #3}}
    }
  }
}

\NewDocumentCommand{\laplaceop}{o m o}{%
  \IfNoValueTF{#1}{%
    \Delta #2 \IfNoValueF{#3}{\left( #3 \right)}
  }{%
    \left.
      \Delta #2
    \right|_{#1\IfNoValueF{#3}{= #3}}
}
}

\NewDocumentCommand{\rintegral}{o o m m}{%
  \int\IfNoValueF{#1}{_{#1}}\IfNoValueF{#2}{^{#2}} #4 \,\diff[1]{#3}%
}

\makeatletter

\newcommand*{\@given}[1]{%
  \nonscript\:#1\vert
  \allowbreak
  \nonscript\:
  \mathopen{}}

\providecommand*{\given}{}

\DeclarePairedDelimiterX{\probparens}[1]{(}{)}{%
  \renewcommand*{\given}{\@given{\delimsize}}%
  #1%
}

\makeatother

\newcommand*{\pmeas}{\mathrm{P}}

\NewDocumentCommand{\prob}{o s O{} m}{%
  \IfValueTF{#1}{#1}{\pmeas}%
  \IfBooleanTF{#2}{%
    \probparens*{#4}
  }{%
    \probparens[#3]{#4}
  }%
}

\NewDocumentCommand{\pdf}{O{p} s O{} m}{%
  #1%
  \IfBooleanTF{#2}{%
    \probparens*{#4}%
  }{%
    \probparens[#3]{#4}%
  }
}

\newcommand*{\rvec}[1]{{\bm{\mathrm{#1}}}}

\newcommand*{\rvf}{\rvec{f}}

\makeatletter

\DeclarePairedDelimiterX{\@condrv}[1]{.}{.}{%
  \renewcommand*{\given}{\@given{\delimsize}}%
  #1}

\NewDocumentCommand{\condrv}{som}{%
  \IfBooleanTF{#1}{%
    \@condrv*{#3}
  }{%
    \IfNoValueTF{#2}{%
      \begingroup%
      \renewcommand*{\given}{\@given{}}%
      #3%
      \endgroup%
    }{%
      \@condrv[#2]{#3}%
    }
  }
}

\makeatother

\NewDocumentCommand{\expectation}{o o m}{%
  \operatorname{\mathbb{E}}\IfNoValueF{#1}{_{#1\IfNoValueF{#2}{\sim #2}}} \left[ #3 \right]
}
\NewDocumentCommand{\covariance}{o o m m}{%
  \operatorname{Cov}\IfNoValueF{#1}{_{#1\IfNoValueF{#2}{\sim #2}}} \left[ #3, #4 \right]
}
\NewDocumentCommand{\variance}{o o m}{%
  \operatorname{\mathbb{V}}\IfNoValueF{#1}{_{#1\IfNoValueF{#2}{\sim #2}}} \left[ #3 \right]
}

\newcommand*{\gaussian}[2]{{\ensuremath{\operatorname{\mathcal{N}}\left(#1, #2\right)}}}
\newcommand*{\gaussianpdf}[3]{%
  {\ensuremath{\operatorname{\mathcal{N}}\left(#1; #2, #3\right)}}%
}

\newcommand*{\tpdf}[4]{%
  {\ensuremath{\operatorname{\mathcal{T}}_{#2}\left(#1; #3, #4\right)}}%
}

\NewDocumentCommand{\LkL}{m m o}{%
  #1 #2 \IfValueTF{#3}{#3}{#1}\dualop
}



%% file: preamble/05_notation.tex
\usepackage{xspace}

\newcommand*{\mdp}{\mathcal{M}}

\newcommand*{\llm}{\pi_\vtheta}
\newcommand*{\prm}{f_\vphi}

\newcommand*{\indicator}[1]{{\mathds{1}[#1]}}
\newcommand*{\concat}[2]{{#1 \oplus #2}}

\newcommand*{\logit}[1]{\operatorname{logit}(#1)}
\newcommand*{\expit}[1]{\operatorname{logit}^{-1}(#1)}
\newcommand*{\dataset}{\mathcal{D}}

%% file: preamble/98_draft.tex
\usepackage[
  obeyDraft,
  textsize=tiny,
  figwidth=0.99\linewidth,
]{todonotes}

%% file: preamble/99_end.tex
\usepackage[
  capitalize,
  nameinlink,
  noabbrev,
]{cleveref}

\makeatletter
\if@cref@capitalise
  \crefname{assumption}{Assumption}{Assumptions}
\else
  \crefname{assumption}{assumption}{assumptions}
\fi
\makeatother

%% file: 00_abstract.tex
While chain-of-thought prompting with Best-of-N (BoN) selection has become popular for mathematical reasoning in large language models (LLMs), its linear structure fails to capture the branching and exploratory nature of complex problem-solving. 
In this work, we propose an adaptive algorithm to maximize process reward model (PRM) scores over the intractable action space, and investigate whether PRM-guided tree search can improve mathematical reasoning by exploring multiple partial solution paths.
Across $23$ diverse mathematical problems using Qwen2.5-Math-7B-Instruct with its associated PRM as a case study, we find that: (1)~PRM-guided tree search shows no statistically significant improvements over BoN despite higher costs, (2)~Monte Carlo tree search and beam search outperform other PRM-guided tree search methods, (3)~PRMs poorly approximate state values and their reliability degrades with reasoning depth, and (4)~PRMs generalize poorly out of distribution. 
This underperformance stems from tree search's greater reliance on unreliable PRM scores, suggesting different reward modeling is necessary before tree search can effectively enhance mathematical reasoning in LLMs.

%% file: 01_intro.tex
\section{Introduction}
\label{sec:intro}
%
Mathematical reasoning involves understanding complex problems, decomposing them into manageable steps, and revisiting intermediate results until reaching a sound solution.
Large language models (LLMs) have shown remarkable capabilities in solving mathematical problems by breaking solutions into reasoning steps through \emph{chain-of-thought} (CoT) prompting \citep{wei2022cot,sprague2025to}.
When combined with process reward models (PRMs) that evaluate individual reasoning steps, Best-of-N (BoN) identifies the most promising CoT from multiple candidates and has become widely adopted \citep{zhang2025lessonsdevelopingprocessreward,yang2024qwen25mathtechnicalreportmathematical,lightman2024lets,uesato2022solvingmathwordproblems}.
However, CoT's linear structure fails to capture the branching nature of mathematical reasoning, where multiple strategies are considered, partial arguments explored, and errors necessitate backtracking \citep{muennighoff2025s1,deepseekai2025deepseekr1incentivizingreasoningcapability}. 
Moreover, restricting PRM evaluation to complete CoTs misses opportunities for dynamic guidance.

The \emph{tree-of-thought} (ToT) framework \citep{yao2023tree} addresses these limitations by exploring multiple partial reasoning paths and enabling revisions using a reward model to assess the correctness of intermediate solutions.
Yet applying ToT with PRMs presents challenges: reasoning trees exhibit intractable branching factors and depth, while PRMs may fail to accurately evaluate intermediate steps \citep{zhang2025lessonsdevelopingprocessreward}.

This work proposes an adaptive algorithm to maximize PRM scores over the intractable action space and empirically investigates whether PRM-guided tree search can improve mathematical reasoning in LLMs.
We evaluate tree search algorithms under varying PRM quality assumptions against BoN across $23$ diverse mathematical problems,  using Qwen2.5-Math-7B-Instruct and its associated PRM as our case study.
Key findings reveal that:~(1) PRM-guided tree search fails to outperform BoN despite higher costs;~(2) Monte Carlo tree search and beam search outperform other PRM-guided tree search methods;~(3) PRMs poorly approximate state values and reliability degrades with reasoning depth, suggesting credit assignment issues; and~(4) PRMs exhibit limited out-of-distribution generalization.
This underperformance stems from tree search's greater reliance on unreliable PRM scores to guide search, whereas BoN evaluates only complete CoTs.
These results highlight the limitations of PRM-guided tree search and BoN, indicating that different reward models may be required for mathematical reasoning.

\paragraph{Limitations.}
This work demonstrates that PRM-guided tree search fails to outperform BoN due to PRM limitations: poor reasoning step value estimation, degraded reliability with reasoning depth, and limited out-of-distribution generalization. 
While our study focuses on a single PRM-model pair, we expect this underperformance to generalize to PRMs exhibiting similar pathologies.
\vspace{-0.45em}

%% file: 02_background.tex
\section{Background}
\label{sec:background}

\subsection{Mathematical reasoning as tree search}
\label{sec:math_reasoning_ts}

We formulate mathematical reasoning as search in a \emph{tree}-structured Markov decision process $\mdp=~(\sS, \sA, r, t)$ where actions $a \in \sA$ are reasoning steps (text ending with an end-of-reasoning-step token), states $s \in \sS \defeq \cup_{i=0}^T \sA^{i}$ are partial reasoning sequences and transitions are deterministic $t(s, a, s') = \indicator{s' = \concat{s}{a}}$ (here $\concat{s}{a}$ denotes string concatenation). 
The root state is the prompt $p$, and $\sT$ denotes terminal states containing predictions in the '\verb|\boxed{x}|' format.
The reward function assigns $r(s) = 1$ if $s \in \sT$ contains the correct solution with valid intermediate reasoning steps, and $r(s) = 0$ otherwise. 
The value function $v(s) = r(s) + \max_{a \in \sA} v(\concat{s}{a})$ indicates whether any continuation from state $s$ leads to a state with reward $1$. 
A LLM $\llm$ defines a policy $a \sim \pdf[\llm]{\cdot \given s}$, while a process reward model $\prm(s)$ estimates $\pdf{v(s) = 1 \given s}$.
Our goal is finding $s$ with $r(s) = 1$ using the LLM and the PRM. 
Since rewards are unavailable during search, the PRM's approximation quality determines the appropriate search strategy. 
We distinguish three scenarios:

\textbf{Scenario 1: PRM as Value Function.} If the PRM correctly ranks actions by their true ${\pdf{v(s') = 1 \given s'=\concat{s}{a}}}$, optimal search recursively selects $a^* = \arg\max_{a \in \sA} \prm(\concat{s}{a})$. 
If actions $\sA$ were practically enumerable, a greedy tree search algorithm would be optimal. 
However, enumerating $a \in \sA$ is intractable and we propose an algorithm in \cref{sec:methods} to adaptively resample actions $a \sim \pdf[\llm]{\cdot \given s}$ until we are confident that we have attained $\max_{a \in \sA} \prm(\concat{s}{a})$.

\textbf{Scenario 2: PRM as Terminal Signal Only.} In this ``worst-case we can still work with'' scenario, the PRM suffers from poor credit assignment, failing to properly estimate $\pdf{v(s) = 1 \given s}$ for intermediate states, yet PRM scores at terminal states still correlate with reward $r$. 
The optimal tree search algorithm returns $s^* \in \arg\max_{s \in \sT} \prm(s)$ among terminal states.

\textbf{Scenario 3: PRM as Noisy Intermediate Signal.} The PRM provides useful but unreliable guidance for intermediate states. 
For example, it may undervalue (i.e., $\prm(s) < \pdf{v(s) = 1 \given s}$) optimal intermediate states leading to high-scoring terminal states, and overvalue (i.e., ${\prm(s) > \pdf{v(s) = 1 \given s}}$) suboptimal intermediate states leading to poor terminal states.
While maximizing PRM scores generally helps to reach valuable terminal nodes, the appropriate search objective remains unclear.

\subsection{Tree search baselines}
\textbf{Best-of-N} samples $N$ chains-of-thought from the LLM policy (i.e., separate root-to-terminal paths with no shared intermediate state) and selects the CoT with the highest aggregated PRM score \citep{zhang2025lessonsdevelopingprocessreward,lightman2024lets}.
Given a prompt $p$, we sample CoT $c_i = (p, \smash{a^{(i)}_1}, \dots, \smash{a^{(i)}_T})$ as $\smash{a^{(i)}_{j+1}} \sim \pdf[\llm]{\cdot \given p \oplus (\oplus_{k=1}^j \smash{a^{(i)}_{k}})}$ for $i=1, \dots, N$ and return $\arg \max_{i \in \set{1, \dots, N}} \Psi(\set{\prm(\smash{a_j^{(i)}})}_{j=1}^T)$ where $\Psi$ aggregates PRM scores.

\textbf{Greedy best-first search (GBFS)} expands the frontier state $s \in \sF$ with highest heuristic value $h(s)$ at each step, where the frontier $\sF$ contains all unexpanded states in the current search tree. Starting from the root, we repeatedly expand $s = \arg\max_{s \in \sF} h(s)$ by sampling $K$ actions from the LLM until reaching a terminal state. We use $h(s) = \prm(s)$ and depth-aware $h(s) = \prm(s) \cdot (M - d(s))$ to favor deeper states, where $M$ is maximum depth and $d(s)$ is the depth of state $s$.

\textbf{Beam search} maintains the top-$N$ states with highest (cumulative) PRM score in a beam $\sB_t$. 
At each step $t$, we sample actions $\smash{a_j^i \sim \pdf[\llm]{\cdot \given s_j}}$ $i=1, \dots, K$ for each state in the current beam $s_j \in \sB_t$, score states $\cup_{j=1}^N \set{s_j \oplus a_j^i}_{i=1}^K$ with the PRM, and keep the $N$ highest-scoring states for the next beam $\sB_{t+1}$. For $N=1$, this reduces to greedy search.

\textbf{Monte Carlo tree search (MCTS)} builds a search tree in four phases: (1) \textbf{Select}: traverse the search tree from root to leaf selecting high-value states and balancing exploration/exploitation; (2) \textbf{Expand}: add children to the selected leaf by sampling actions from $\llm$; (3) \textbf{Rollout}: run LLM policy from the new state to a terminal state; (4) \textbf{Backpropagate}: update visit counts and average PRM scores of terminal states along the path back to the root. Repeat until computational budget is depleted.
\vspace{-0.5em}

%% file: 03_methods.tex
\section{Methods}
\label{sec:methods}
In this section, we propose an adaptive algorithm to solve the intractable optimization problem $a = \arg\max_{a \in \sA} \prm(\concat{s}{a})$ (due to the unenumerable $\card{\sA}$) from Scenario 1 (see \cref{sec:math_reasoning_ts}). 
We formulate this optimization problem as a stopping problem: when should we stop sampling actions from the policy and commit to the action with highest observed $\prm(\concat{s}{a})$?

\textbf{Maximizing over the intractable $\sA$ using Gittin's indices.}
At each state $s$, we sample independent actions $a_i \sim \pdf[\llm]{\cdot \given s}$ and evaluate PRM scores $f_i = \prm(\concat{s}{a_i})$. 
We must then decide whether to \textbf{stop} and commit to the current maximum \emph{observed} PRM score $m = \max_i f_i$ (payoff $m$), or \textbf{sample} again at cost $c$ to improve the current estimate $m$ for expected payoff $\expectation[f][\pdf{f \given s}]{\max(m, f)} - c$ (since we select the largest value among $f \sim \pdf{\cdot \given s}$ and $m$ and incur a cost $c$).
This is an instance of the Pandora's box problem \citep{weitzman1979pandora}, whose optimal strategy \textbf{samples} if $\expectation[f][\pdf{f \given s}]{\max(m, f)} - c > m$ and otherwise \textbf{stops}.
This involves computing a Gittin's index $m^*$ satisfying $\expectation[f][\pdf{f \given s}]{\max(0, f-m^*)} = c$, then sampling if $m^* > m$ and stopping otherwise.
However, $\pdf{f \given s}$ is intractable and estimating it requires the LLM samples we are trying to acquire sparingly.
This motivates a strategy based on surrogate modeling and posterior inference inspired by \citet{xie2024costaware} which we discuss next.

\textbf{Bayesian surrogate approximation.}
We approximate $\pdf{f \given s}$ using a logit-Normal surrogate model $\pdf[q]{f \given s, \vpsi}$ with parameters $\vpsi$.
Specifically, we encode prior beliefs in $\pdf{\vpsi}$, update the posterior $\pdf{\vpsi \given \dataset} \propto \pdf[q]{\dataset \given s, \vpsi} \pdf{\vpsi}$ with observed PRM scores $\dataset = \set{f_i | f_i \sim \pdf{f \given s}}$, and use the posterior predictive $\pdf[q]{f \given s, \dataset} = \int \pdf[q]{f \given s, \vpsi} \pdf{\vpsi \given \dataset} d\vpsi$ to approximate $\pdf{f \given s}$. 
We then compute the Gittin's index $m^*$ by solving $\expectation[f][\pdf[q]{f \given s, \dataset}]{\max(0, f - m)} = c$ under posterior beliefs $\pdf[q]{f \given s, \dataset}$ rather than $\pdf{f \given s}$.
The left-hand side is the expected improvement over $m$ \citep{jones1998EI}, a standard Bayesian optimization acquisition function \citep{garnett_bayesoptbook_2023}.
The Gittin's index represents the threshold where expected improvement equals cost $c$, thus smaller $c$ induces more exploration. More details in \cref{sec:add_methods}.
\vspace{-0.5em}

%% file: 04_results.tex
\section{Results}
\label{sec:results}
\subsection{Experimental setup}
\paragraph{LLM \& PRM.} We use the Qwen2.5-Math-7B-Instruct \citep{yang2024qwen25mathtechnicalreportmathematical} LLM with the recommended prompting strategy and sampling parameters, and the Qwen2.5-Math-PRM-7B process reward model \citep{zhang2025lessonsdevelopingprocessreward}.
\paragraph{Problems \& metrics.} We evaluate on $22$ mathematical reasoning problems from \citet{yang2024qwen25mathtechnicalreportmathematical} and AIME 2025 \citep{balunovic_srimatharena_2025}. 
We report mean accuracy and rank across problems with standard errors.
To address concerns about high variance in LLM evaluation \citep{hochlehnert2025soberlookprogresslanguage}, we test statistical significance between the top-performing method and all others using Wilcoxon signed-rank tests (insignificant if $p > 0.05$).
\paragraph{Tree search methods.} We compare Best-of-N (\texttt{BoN}) with $N=8$ chain-of-thoughts using last, minimum, average, product, maximum and sum aggregation functions $\Psi$; the proportion of answers containing at least one correct prediction among $N=8$ CoTs (\texttt{PASS@N}); majority voting among predictions of $N=8$ CoTs (\texttt{MAJ@N}); beam search with beam size $N=1$ expanding the state with highest PRM value from $K$ policy samples (\texttt{Greedy@K}); beam search with beam size $N=4$ from $K=6$ policy samples maximizing instantaneous (V) or cumulative (CV) PRM scores (\texttt{Beam@N}); greedy best-first search with $K=8$ policy samples (\texttt{GBFS@K}); depth-aware GBFS (\texttt{GBFS\_DA@K}; see \cref{sec:background}); Monte Carlo tree search with $K=8$ policy samples (\texttt{MCTS@K}) and our proposed method from \cref{sec:methods} with constant and linear cost schedules to allow more exploration early in the search when the remaining sampling budget is large (\texttt{Gittins@cost}; more details in \cref{sec:add_methods}).
\subsection{Findings}
%
\begin{wraptable}{r}{0.47\textwidth}
\vspace{-1.25em}
\centering
\caption{\textbf{Method mean accuracy and rank with standard errors across problems.} We bold results which are not significantly worse than the best ($p > 0.05$).}
\label{tab:acc_rank}
\resizebox{0.47\textwidth}{!}{
\begin{tabular}{lcc}
\toprule
\textsc{Method} & \textsc{Accuracy (p-value)} & \textsc{Rank (p-value)} \\
\midrule
\texttt{PASS@8} & \textbf{79.8 $\pm$ 4.7 (N/A)\hphantom{0}} & N/A \\
\midrule
\texttt{MAJ@8} & 71.4 $\pm$ 5.1 (0.010) & 4.43 $\pm$ 0.51 (0.022) \\
\midrule
\texttt{BoN\_Last@8} & \textbf{72.7 $\pm$ 5.1 (N/A)\hphantom{0}} & \textbf{3.13 $\pm$ 0.38 (N/A)\hphantom{0}} \\
\texttt{BoN\_Avg@8} & \textbf{72.1 $\pm$ 5.0 (0.444)} & \textbf{3.22 $\pm$ 0.31 (0.787)} \\
\texttt{BoN\_Min@8} & \textbf{72.2 $\pm$ 5.0 (0.711)} & \textbf{3.26 $\pm$ 0.32 (0.608)} \\
\texttt{BoN\_Prod@8} & \textbf{72.0 $\pm$ 5.0 (0.408)} & \textbf{3.26 $\pm$ 0.40 (0.795)} \\
\texttt{BoN\_Sum@8} & 67.6 $\pm$ 5.3 (0.000) & 7.30 $\pm$ 0.72 (0.000) \\
\texttt{BoN\_Max@8} & 68.8 $\pm$ 5.4 (0.000) & 7.35 $\pm$ 0.69 (0.000) \\
\midrule
\texttt{Greedy@6} & 71.6 $\pm$ 5.0 (0.043) & 5.74 $\pm$ 0.76 (0.003) \\
\texttt{Greedy@20} & 71.2 $\pm$ 5.0 (0.039) & 5.09 $\pm$ 0.55 (0.009) \\
\midrule
\texttt{Beam@4 (V)} & \textbf{71.8 $\pm$ 4.9 (0.126)} & \textbf{3.83 $\pm$ 0.42 (0.236)} \\
\texttt{Beam@4 (CV)} & \textbf{71.9 $\pm$ 4.8 (0.189)} & \textbf{4.00 $\pm$ 0.49 (0.221)} \\
\midrule
\texttt{GBFS@8} & 46.0 $\pm$ 4.1 (0.000) & 10.09 $\pm$ 0.71 (0.000) \\
\texttt{GBFS\_DA@8} & 48.1 $\pm$ 4.6 (0.000) & 10.00 $\pm$ 0.65 (0.000) \\
\midrule
\texttt{MCTS@8} & \textbf{71.2 $\pm$ 5.0 (0.987)} & \textbf{3.26 $\pm$ 0.69 (0.856)} \\
\midrule 
\texttt{Gittins@0.05} (ours) & 70.5 $\pm$ 5.2 (0.012) & 5.96 $\pm$ 0.59 (0.001) \\
\texttt{Gittins@linear} (ours) & 71.4 $\pm$ 5.1 (0.013) & 4.74 $\pm$ 0.56 (0.011) \\
\bottomrule
\end{tabular}
}
\vspace{-1em}
\end{wraptable} 
\paragraph{1. PRM-guided tree search methods do not outperform Best-of-N despite higher costs.}
Best-of-N using terminal PRM scores (\texttt{BoN\_Last@8}) achieves the highest mean rank and accuracy (see \cref{tab:acc_rank,tab:acc_all_datasets}). 
Among Best-of-N variants, average, minimum, and product aggregations perform comparably without significant differences. 
MCTS and beam search also show no significant performance degradation compared to the best method.
However, tree search methods incur substantially higher computational costs, generating considerably more tokens (see Generated token in \cref{tab:acc_rank_extended}). Despite this increased cost, their final solutions contain fewer reasoning steps and tokens than Best-of-N solutions (see Reasoning steps and Out Tokens in \cref{tab:acc_rank_extended}). 
Except for GBFS, tree search methods reach approximately as many terminal states as Best-of-N (see Terminal states in \cref{tab:acc_rank_extended}).
%
\begin{wraptable}{r}{0.375\textwidth}
\vspace{-1.2em}
\centering
\caption{\textbf{Method mean accuracy and rank with standard errors across problems for tree search methods.} We bold results which are not significantly worse than the best ($p > 0.05$).}
\label{tab:acc_rank_ts}
\resizebox{0.375\textwidth}{!}{
\begin{tabular}{lcc}
\toprule
\textsc{Method} & \textsc{Accuracy (p-value)} & \textsc{Rank (p-value)} \\
\midrule
\texttt{Greedy@6} & \textbf{71.6 $\pm$ 5.0 (0.498)} & 3.78 $\pm$ 0.41 (0.021) \\
\texttt{Greedy@20} & 71.2 $\pm$ 5.0 (0.019) & 3.43 $\pm$ 0.27 (0.032) \\
\midrule
\texttt{Beam@4 (V)} & \textbf{71.8 $\pm$ 4.9 (0.601)} & \textbf{2.52 $\pm$ 0.20 (0.232)} \\
\texttt{Beam@4 (CV)} & \textbf{71.9 $\pm$ 4.8 (N/A)\hphantom{0}} & \textbf{2.57 $\pm$ 0.24 (0.286)} \\
\midrule
\texttt{GBFS@8} & 46.0 $\pm$ 4.1 (0.000) & 6.70 $\pm$ 0.34 (0.000) \\
\texttt{GBFS\_DA@8} & 48.1 $\pm$ 4.6 (0.000) & 6.48 $\pm$ 0.30 (0.000) \\
\midrule
\texttt{MCTS@8} & \textbf{71.2 $\pm$ 5.0 (0.332)} & \textbf{2.26 $\pm$ 0.41 (N/A)\hphantom{0}} \\
\midrule
\texttt{Gittins@0.05} (ours) & 70.5 $\pm$ 5.2 (0.019) & 4.00 $\pm$ 0.35 (0.006) \\
\texttt{Gittins@linear} (ours) & \textbf{71.4 $\pm$ 5.1 (0.398)} & 3.09 $\pm$ 0.30 (0.048) \\
\bottomrule
\end{tabular}
}
\vspace{-2em}
\end{wraptable}
\paragraph{2. MCTS and beam search perform best among PRM-guided tree search methods.}
\texttt{MCTS@8} achieves the highest mean rank among tree search methods, with beam search variants (\texttt{Beam@4 (V)} and \texttt{Beam@4 (CV)}) performing comparably without significant differences (${p > 0.05}$, see \cref{tab:acc_rank_ts}). 
For mean accuracy, beam search maximizing the cumulative PRM values performs best, followed by \texttt{Beam@4 (V)}, \texttt{Greedy@6}, \texttt{Gittins@linear} and \texttt{MCTS@8} with no significant performance gaps.
\begin{wrapfigure}{r}{0.4\textwidth}
\vspace{-1.05em}
\centering
\includegraphics[width=0.4\textwidth]{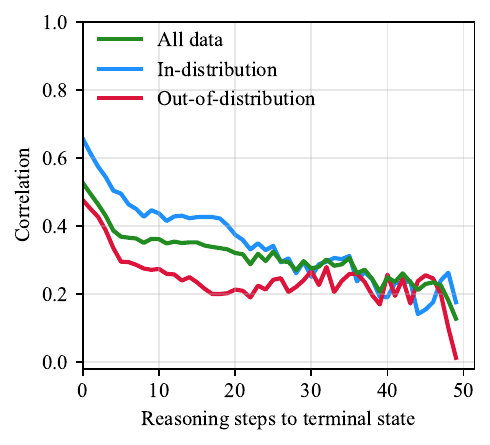}
\caption{\textbf{Correlation of prediction correctness with PRM scores.} Correlation decreases with increasing distance in reasoning steps from terminal states.}
\label{fig:correlation_by_reasoning_step}
\vspace{-2.5em}
\end{wrapfigure}
\paragraph{3. The PRM poorly approximates ${\pdf{v(s) = 1 \given s}}$ and reliability degrades with reasoning depth, limiting tree search effectiveness.}
Methods assuming the PRM accurately estimates the value across all states (\texttt{Greedy@K} and \texttt{Gittins@cost}, Scenario 1 in \cref{sec:math_reasoning_ts}) perform significantly worse than Best-of-N and other tree search methods (\cref{tab:acc_rank,tab:acc_rank_ts}). 
Increasing policy samples ($K=6$ to $K=20$) or adaptive sampling (\texttt{Gittins@cost}) does not improve performance.
This suggests either the LLM policy cannot generate high-scoring states or the PRM cannot identify them. 
Since \texttt{PASS@K} significantly outperforms Best-of-N, the LLM policy does generate correct solutions but the PRM fails to rank them highly, providing evidence that the PRM poorly approximates $\pdf{v(s) = 1 \given s}$.

Further analysis shows that point-biserial correlation of prediction correctness with PRM scores is initially high (${\approx 0.5}$) near terminal states but deteriorates significantly for early reasoning steps
($\approx 0.37$ at $10$ steps from termination; see All data in \cref{fig:correlation_by_reasoning_step}). 
These findings suggest credit assignment issues in the offline reinforcement learning of the PRM.
Moreover, this pattern explains why \texttt{MCTS@8}, which relies exclusively on terminal PRM scores, achieves the best average rank among tree search methods, and \texttt{Beam@8} performs best in accuracy by tolerating locally suboptimal steps that lead to higher-scoring future states. 
This suggests that the PRM operates between Scenarios 2 and 3: terminal scores are most reliable, but intermediate scores provide useful yet unreliable guidance.
\paragraph{4. The PRM shows limited out-of-distribution generalization.} 
Correlation between PRM scores and correctness is consistently higher on in-distribution (ID) problems on which the PRM is trained (GSM8K and MATH) than out-of-distribution (OOD) tasks (others problems; \cref{fig:correlation_prm_correctness,tab:corr_all_datasets}). 
This generalization gap persists across most reasoning steps: correlation of prediction correctness with PRM scores on ID problems is considerably larger than on OOD tasks until $\approx 30$ steps to termination, after which both ID and OOD performance converge to similarly low correlation levels (see ID and OOD in \cref{fig:correlation_by_reasoning_step}).
This limited generalization further constrains the practical utility of PRM-guided tree search across diverse mathematical domains.

%% file: 05_related_work.tex
\section{Related work}

\paragraph{Tree search with LLM self-evaluation} 
Several works apply greedy best-first search \citep{yao2023tree}, Monte Carlo tree search \citep{zhang2024accessinggpt4levelmathematical, hao2023reasoning} and beam search \citep{xie2023selfevaluation} to reasoning using the LLM policy model itself for state evaluation. 
Unlike these approaches, we use a process reward model trained by offline reinforcement learning on mathematics tasks to guide search.

\paragraph{PRM-guided tree search} 
\citet{zhang2025lessonsdevelopingprocessreward} evaluate PRM aggregation methods and greedy search, finding greedy search generally inferior to Best-of-N. 
Our study extends this analysis with a more comprehensive evaluation, including additional tree search methods (MCTS, GBFS, beam search) and more than three times as many problems ($23$ vs. $7$).
Our findings challenge some of theirs results, notably that last-token aggregation outperforms product aggregation on average.
More importantly, we provide evidence that PRM reliability degrades with reasoning depth, a key finding that explains why tree search methods fail to outperform Best-of-N. 

\paragraph{Tree search for LLM training} 
Recent work has used MCTS within reinforcement learning pipelines to improve both language models and PRMs. \citet{zhang2024restmcts,wan2024alphats,xie2024mctsllm,guan2025rstarmath} employ MCTS under PRM supervision to generate high-scoring reasoning chains subsequently used to fine-tune both the policy and PRM through self-training. 
In contrast, our work focuses on pretrained LLM and PRM models without fine-tuning.

%% file: 06_conclusion.tex
\section{Conclusion}
\label{sec:conclusion}

We proposed an adaptive algorithm to maximize PRM scores over the intractable action space, and empirically investigated PRM-guided tree search across $23$ mathematical reasoning problems using Qwen2.5-Math-7B-Instruct and its associated PRM. Our findings show that PRM-guided tree search methods fail to outperform Best-of-N despite higher costs, with MCTS and beam search proving most effective among PRM-guided tree search approaches. 
We identify the underlying causes: PRMs poorly approximate state values, become less reliable with reasoning depth indicating credit assignment issues, and exhibit limited out-of-distribution generalization restricting their broader applicability. 
Tree search's underperformance stems from its reliance on these unreliable intermediate-step PRM scores to guide search, whereas BoN evaluates only complete CoTs.
These results highlight the limitations of both PRM-guided tree search and BoN, revealing that current PRMs lack sufficient accuracy to guide dynamic mathematical reasoning and suggesting that different reward models may be required.

%% file: 99_ack.tex
GP is supported by the Canada CIFAR AI Chair program. The authors are grateful to Johannes Zenn for many insightful discussions that helped shape the direction of this project, and to both Johannes Zenn and Arsen Sheverdin for their careful reading of the manuscript and valuable feedback that improved its clarity and presentation. The computational resources used in this work were provided, in part, by the Province of Ontario, the Government of Canada through CIFAR, and by companies sponsoring the Vector Institute\footnote{\url{[https://vectorinstitute.ai/partnerships/}}.

%% file: 07_appendix.tex
\appendix

\input{appendix/A_methods}
\input{appendix/B_experimental_setup}

\input{appendix/C_experimental_results}

%% file: appendix/A_methods.tex
\section{Additional method details}
\label{sec:add_methods}

We here provide additional details on the method presented in \cref{sec:methods}.

\paragraph{Maximizing over the intractable $\sA$ using Gittin's indices.}

At each state $s$, we sample independent actions from the policy $a \sim \pdf[\llm]{\cdot \given s}$ and evaluate $f = \prm(\concat{s}{a})$, which induces the distribution
\begin{equation}\label{eq:prm_dist}
    \pdf{\rvf = f \given s} = \sum_{a \in \sA} \mathds{1}(f = \prm(\concat{s}{a})) \pdf[\llm]{a \given s}.
\end{equation}
After sampling $i$ actions from the policy, let $m_i = \max_{j \leq i} f_j$ denote the maximum PRM score observed so far.
We face a choice between two actions:
\begin{enumerate}
    \item \textbf{Sample} a new action $a_{i+1} \sim \pdf[\pi]{\cdot \given s}$, observe $f_{i+1} = \prm(\concat{s}{a_{i+1}})$, and incur cost $c$
    \item \textbf{Stop} and commit to the current best action with score $m_i$
\end{enumerate}
The expected payoff for sampling is $\expectation[f][\pdf{f \given s}]{\max(m_i, f)} - c$, while stopping yields payoff $m_i$. The optimal policy samples when the sampling payoff exceeds the stopping payoff:
\begin{align}
    \expectation[f][\pdf{f \given s}]{\max(m_i, f)} - c &> m_i \\
    \Leftrightarrow \quad \expectation[f][\pdf{f \given s}]{\max(0, f - m_i)} &> c
\end{align}
The Gittins index $m_i^*$ is defined as the unique solution to
\begin{equation}
    \expectation[f][\pdf{f \given s}]{\max(0, f-m_i^*)} = c
\end{equation}
The optimal policy samples if $m_i^* > m_i$ and stops otherwise \citep{weitzman1979pandora}.
However, computing this expectation requires knowledge of $\pdf{f \given s}$ which is intractable due to the summation over the action space $\sA$ (see \cref{eq:prm_dist}). 
Since estimating this distribution would require the very LLM samples we aim to collect sparingly, we develop a Bayesian surrogate approach inspired by \citet{xie2024costaware} which we discuss next.

\paragraph{Bayesian surrogate modeling}

We approximate $\pdf{f \given s}$ using a surrogate model $\pdf[q]{f \given s, \vpsi}$ with parameters $\vpsi$.
Specifically, we encode our prior beliefs in $\pdf{\vpsi}$, then update the posterior $\pdf{\vpsi \given \dataset} \propto \pdf[q]{\dataset \given s, \vpsi} \pdf{\vpsi}$ with observations $\dataset = \set{f_i | f_i \sim \pdf{f \given s}}$ before using the posterior predictive $\pdf[q]{f \given s, \dataset} = \int \pdf[q]{f \given s, \vpsi} \pdf{\vpsi \given \dataset} d\vpsi$ to approximate $\pdf{f \given s}$. 
We then compute the Gittin's index $m_i^*$ under posterior beliefs $\pdf[q]{f \given s, \dataset}$ rather than $\pdf{f \given s}$ by solving 
\begin{equation} \label{eq:pandora_gittins_idx}
    \expectation[f][\pdf[q]{f \given s, \dataset}]{\max(0, f - m_i)} = c
\end{equation}
The left-hand side of \cref{eq:pandora_gittins_idx} is the expected improvement over the current maximum $m$ \citep{jones1998EI}, a standard Bayesian optimization acquisition function \citep{garnett_bayesoptbook_2023}.
The Gittin's index $m^*$ represents the threshold where expected improvement equals cost $c$, thus smaller $c$ induces more exploration. More details in \cref{alg:pandora}.

\paragraph{Adaptive cost scheduling.}
To balance exploration and exploitation over the search horizon, we employ time-varying costs:
\begin{equation}
    c(n) = c_1 + (c_2 - c_1) \times \nicefrac{n}{B}
\end{equation}
where $c_1 < c_2$ are initial and final costs, $B$ is the total sampling budget, and $n$ is the current sample count.
This schedule promotes exploration early during search when the remaining sampling budget is large and exploitation as resources diminish.

\begin{algorithm}[t]
\caption{Adaptive PRM-guided tree search using Gittin's indices and surrogate modeling.}
\label{alg:pandora}
\begin{algorithmic}[1]
\Function{Gittins}{$\prm$, $\llm$, $\sT$, $s$, $K$, $c$}
  \Repeat 
  \State $\dataset \gets \emptyset$
  \Repeat
    \State $\dataset \gets \dataset \cup \set{(s_i, \prm(s_i)) \,|\, a_i \sim \pdf[\llm]{\cdot \given s}, s_i = \concat{s}{a_i}}_{i=1}^K$ \Comment{Sample from LLM}
    \State $m \gets \max_{(s, f) \in \dataset} f$  \Comment{Update current observed maximum}
    \State Compute posterior predictive $\pdf[q]{f \given s, \dataset}$ \Comment{Update posterior beliefs}
    \State Compute Gittin's index $m^*$ by solving $\expectation[\pdf[q]{f \given \dataset}]{\max(0, f-m^*)} = c$ using bisection
    \Until{$m^* > m$}
    \State $s \gets \max_{(s, f) \in \dataset} f$ \Comment{Update current state}
    \Until{$s \in \sT$}
  \State \textbf{return} $s$
\EndFunction
\end{algorithmic}
\end{algorithm}

\paragraph{Logit-Normal surrogate model}

Let $\dataset_n = \set{f_i}_{i=1}^n$ where $f_i \in [0,1]$ be a collection of observations from $\pdf{\rvf = f \given s}$ at a given state $s$.
We consider a logit-Normal likelihood model for our observations i.e. 
\begin{equation}
    \pdf[q]{f \given s} = \gaussianpdf{\logit{f}}{\mu}{\sigma^2} \frac{1}{f(1-f)}
\end{equation}
where $\frac{1}{f(1-f)}$ adjusts for the change of variable.
We then specify a Normal-inverse-Gamma prior on the likelihood parameters $\mu$ and $\sigma^2$ i.e.
\begin{equation}
    \pdf[q]{\mu, \sigma^2} = \gaussian{m_0}{v_0 \sigma^2} \pdf[\mathcal{IG}]{\alpha_0, \beta_0}
\end{equation}
The model is conjugate and both the posterior and the predictive posterior are available in closed form.
The prior is chosen to yield a maximally uniform predictive prior.

The prior predictive is 
\begin{equation}
    \pdf[q]{f \given \mu, \sigma^2} = \int \pdf[q]{r \given \mu, \sigma^2} \pdf[q]{\mu, \sigma^2} d\mu d\sigma^2 = \tpdf{f}{2 a_0}{m_0}{\frac{b_0(1+v_0)}{a_0}} \frac{1}{f(1-f)}
\end{equation}

The posterior is then 
\begin{align}
    \pdf[q]{\mu, \sigma^2 \given \dataset} &= \frac{\prod_{i=1}^n \pdf[q]{f_i \given s} \pdf[q]{\mu, \sigma^2}}{\int \prod_{i=1}^n \pdf[q]{f_i \given s} \pdf[q]{\mu, \sigma^2} d\mu d\sigma^2} \\
    &= \frac{C \prod_{i=1}^n \gaussianpdf{\logit{f_i}}{\mu}{\sigma^2} \pdf[q]{\mu, \sigma^2}}{C \int \prod_{i=1}^n \gaussianpdf{\logit{f_i}}{\mu}{\sigma^2} \pdf[q]{\mu, \sigma^2} d\mu d\sigma^2} \\
    &= \frac{\prod_{i=1}^n \gaussianpdf{\logit{f_i}}{\mu}{\sigma^2} \pdf[q]{\mu, \sigma^2}}{\int \prod_{i=1}^n \gaussianpdf{\logit{f_i}}{\mu}{\sigma^2} \pdf[q]{\mu, \sigma^2} d\mu d\sigma^2} \\
    &= \gaussian{m_n}{v_n \sigma^2} \pdf[\mathcal{IG}]{\alpha_n, \beta_n}
\end{align}
where $C = \prod_{i=1}^n\frac{1}{f_i(1-f_i)}$ , $v_n^{-1} = v_0^{-1} + n$, $m_n = v_n^{-1} (v_0^{-1} m_0 + \sum_{i=1}^n \logit{f_i})$, $a_n = a_0 + n / 2$ and $b_n = b_0 + \frac{1}{2}[m_0^2v_0^{-1} + \sum_{i=1}^n \logit{f_i}^2 - m_n^2 v_n^{-1}]$.

The predictive posterior is 
\begin{equation}
    \pdf[q]{f \given \dataset} = \int \pdf[q]{f \given \mu, \sigma^2} \pdf[q]{\mu, \sigma^2 \given \dataset} d\mu d\sigma^2 = \tpdf{f}{2 a_n}{m_n}{\frac{b_n(1+v_n)}{a_n}} \frac{1}{f(1-f)}
\end{equation}

Furthermore, the marginal likelihood has an analytical formulation: 
\begin{align}
    \pdf[q]{\dataset \given m_0, v_0, a_0, b_0} &= \int \prod_{i=1}^n \pdf[q]{f_i \given s} \pdf[q]{\mu, \sigma^2} d\mu d\sigma^2 \\
     &= \left[ \prod_{i=1}^n \frac{1}{f_i(1-f_i)} \right] \left[ \int \prod_{i=1}^n \pdf[q]{\logit{f_i} \given s} \pdf[q]{\mu, \sigma^2} d\mu d\sigma^2\right] \\
     &= C \left[ \frac{v_n^{1/2} b_0^{a_0} \Gamma(a_n)}{v_0^{1/2} b_n^{a_n} \Gamma(a_0) \pi^{n/2}2^n} \right] 
\end{align}
where $C = \prod_{i=1}^n\frac{1}{f_i(1-f_i)}$.

\paragraph{Implementation details}

We estimate the expectation $\expectation[q]{\max(0, f-m)}$ using quadrature reparameterizing the integral to logit-space
\begin{align}
    \expectation[q]{\max(0, f-m)} &= \int_m^1 (f-m) \pdf[q]{f \given \dataset} df \\
                                &= \int_{\logit{m}}^{\logit{1}} (\expit{l}-m) \pdf[q]{l \given \dataset} dl
\end{align}
where $l = \logit{f}$. 
We solve for the Gittin's index $m^*$ using bissection search as done in \citet{xie2024costaware}.
Since the observations $f$ take values in $[0,1]$, we shrink them using $f_i = \epsilon + (1 - 2\epsilon) \prm(s_i)$ to avoid singularities at $0$ and $1$. 

%% file: appendix/B_experimental_setup.tex
\section{Additional experimental setup details}
\label{sec:add_experimental_setup}

\paragraph{Models and hyperparameters.} We use Qwen2.5-Math-7B-Instruct \citep{qwen25} as our base language model, where actions $a \in \mathcal{A}$ correspond to text sequences terminated by '\verb|\n\n|'. 
For process reward modeling, we use Qwen2.5-Math-PRM-7B trained on MATH and GSM8K problems \citep{zhang2025lessonsdevelopingprocessreward}. 
Following the recommended configuration \citep{qwen25}, we set temperature=0.7, top\_p=0.8, and repetition\_penalty=1.05 for text generation.

\paragraph{Implementation.} Gittins indices are computed via bisection search and  numerical integration performed using SciPy \citep{pauli2020scipy}. 
Tree structures are managed through NetworkX \citep{networkx}. 
Models are served using vLLM \citep{kwon2023efficient} with our inference pipeline adapted from vector-inference \citep{vectorinference2025}.
Evaluation follows the protocol established by \citet{yang2024qwen25mathtechnicalreportmathematical} using their official codebase.\footnote{\url{https://github.com/QwenLM/Qwen2.5-Math}}

\paragraph{Computational resources.} All experiments are conducted on NVIDIA RTX 6000 and A40 GPUs, with each model deployed on a single GPU.

%% file: appendix/C_experimental_results.tex
\section{Additional experimental results}
\label{sec:add_experimental_results}

\begin{table}[htbp]
\centering
\caption{
\textbf{Performance metrics across 23 problems: mean accuracy with standard-error, mean rank with standard-error, reasoning steps, generated tokens (Gen.), output tokens (Out.), and terminal state coverage.} Bold entries indicate no significant difference from the best method ($p > 0.05$). Best-of-N with terminal PRM scores (\texttt{BoN\_Last@8}) perform best, while tree search methods (MCTS, beam search) match accuracy but require substantially higher computational cost.}
\label{tab:acc_rank_extended}
\medskip  
\resizebox{\textwidth}{!}{
\begin{tabular}{lcccccc}
\toprule
\textsc{Method} & \textsc{Accuracy} & \textsc{Rank} & \textsc{Reasoning steps} & \textsc{Gen. Tokens ($\times 10^7$)} & \textsc{Out. Tokens ($\times 10^7$)} & \textsc{Terminal State (\%)} \\
\midrule
\texttt{PASS@8} & \textbf{79.8 $\pm$ 4.7 (N/A)\hphantom{0}} & N/A & 9.7 & 8.2 & 8.2 & 98.2 \\
\midrule
\texttt{MAJ@8} & 71.4 $\pm$ 5.1 (0.010) & 4.43 $\pm$ 0.51 (0.022) & 9.7 & 8.2 & 8.2 & 98.2 \\
\texttt{BoN\_Last@8} & \textbf{72.7 $\pm$ 5.1 (N/A)\hphantom{0}} & \textbf{3.13 $\pm$ 0.38 (N/A)\hphantom{0}} & 9.7 & 8.2 & 8.2 & 98.2 \\
\texttt{BoN\_Avg@8} & \textbf{72.1 $\pm$ 5.0 (0.444)} & \textbf{3.22 $\pm$ 0.31 (0.787)} & 9.7 & 8.2 & 8.2 & 98.2 \\
\texttt{BoN\_Min@8} & \textbf{72.2 $\pm$ 5.0 (0.711)} & \textbf{3.26 $\pm$ 0.32 (0.608)} & 9.7 & 8.2 & 8.2 & 98.2 \\
\texttt{BoN\_Prod@8} & \textbf{72.0 $\pm$ 5.0 (0.408)} & \textbf{3.26 $\pm$ 0.40 (0.795)} & 9.7 & 8.2 & 8.2 & 98.2 \\
\texttt{BoN\_Sum@8} & 67.6 $\pm$ 5.3 (0.000) & 7.30 $\pm$ 0.72 (0.000) & 9.7 & 8.2 & 8.2 & 98.2 \\
\midrule
\texttt{BoN\_Max@8} & 68.8 $\pm$ 5.4 (0.000) & 7.35 $\pm$ 0.69 (0.000) & 9.7 & 8.2 & 8.2 & 98.2 \\
\midrule
\texttt{Greedy@6} & 71.6 $\pm$ 5.0 (0.043) & 5.74 $\pm$ 0.76 (0.003) & 8.9 & 5.6 & 0.9 & 97.9 \\
\texttt{Greedy@20} & 71.2 $\pm$ 5.0 (0.039) & 5.09 $\pm$ 0.55 (0.009) & 8.8 & 18.7 & 0.9 & 98.2 \\
\midrule
\texttt{Beam@4 (V)} & \textbf{71.8 $\pm$ 4.9 (0.126)} & \textbf{3.83 $\pm$ 0.42 (0.236)} & 9.2 & 20.2 & 0.9 & 98.7 \\
\texttt{Beam@4 (CV)} & \textbf{71.9 $\pm$ 4.8 (0.189)} & \textbf{4.00 $\pm$ 0.49 (0.221)} & 9.4 & 21.0 & 0.9 & 96.5 \\
\midrule
\texttt{GBFS@8} & 46.0 $\pm$ 4.1 (0.000) & 10.09 $\pm$ 0.71 (0.000) & 5.3 & 65.2 & 0.5 & 52.7 \\
\texttt{GBFS\_DA@8} & 48.1 $\pm$ 4.6 (0.000) & 10.00 $\pm$ 0.65 (0.000) & 5.8 & 68.0 & 0.5 & 57.1 \\
\midrule
\texttt{MCTS} & \textbf{71.2 $\pm$ 5.0 (0.987)} & \textbf{3.26 $\pm$ 0.69 (0.856)} & 8.8 & 89.4 & 0.7 & 97.8 \\
\midrule
\texttt{Gittins@0.05} (ours) & 70.5 $\pm$ 5.2 (0.012) & 5.96 $\pm$ 0.59 (0.001) & 9.5 & 9.0 & 0.9 & 98.3 \\
\texttt{Gittins@linear} (ours) & 71.4 $\pm$ 5.1 (0.013) & 4.74 $\pm$ 0.56 (0.011) & 9.5 & 14.3 & 0.9 & 98.5 \\
\bottomrule
\end{tabular}
}
\end{table}

\begin{table}[htbp]
\centering
\caption{
\textbf{Accuracy by dataset with means and standard errors.} Bold indicates best performance per dataset. Bottom rows show overall means and win counts across problems.}
\medskip  
\label{tab:acc_all_datasets}
\resizebox{\textwidth}{!}{
\begin{tabular}{lccccccccccccccccc}
\toprule
\textsc{Dataset} & \texttt{PASS@8} & \texttt{MAJ@8} & \texttt{BoN\_Last@8} & \texttt{BoN\_Avg@8} & \texttt{BoN\_Min@8} & \texttt{BoN\_Prod@8} & \texttt{BoN\_Sum@8} & \texttt{BoN\_Max@8} & \texttt{Greedy@6} & \texttt{Greedy@20} & \texttt{Gittins@0.05} (ours) & \texttt{Gittins@linear} (ours) & \texttt{Beam@4 (V)} & \texttt{Beam@4 (CV)} & \texttt{GBFS@8} & \texttt{GBFS\_DA@8} & \texttt{MCTS@8} \\
\midrule
\textsc{aime25} & 20.0 & 13.3 & 13.3 & 16.7 & 16.7 & 20.0 & 13.3 & 6.7 & 16.7 & 10.0 & 13.3 & 16.7 & 20.0 & 16.7 & 6.7 & 3.3 & \textbf{23.3} \\
\textsc{aime24} & 23.3 & 16.7 & 16.7 & 20.0 & 20.0 & 20.0 & 16.7 & 10.0 & 16.7 & 26.7 & 16.7 & 13.3 & 16.7 & \textbf{30.0} & 6.7 & 13.3 & 26.7 \\
\textsc{amc23} & 82.5 & 60.0 & \textbf{70.0} & 67.5 & \textbf{70.0} & \textbf{70.0} & 62.5 & 62.5 & 62.5 & 62.5 & 55.0 & \textbf{70.0} & 67.5 & 65.0 & 40.0 & 37.5 & 65.0 \\
\textsc{sat\_math} & 100.0 & 96.9 & \textbf{100.0} & \textbf{100.0} & \textbf{100.0} & \textbf{100.0} & 93.8 & 93.8 & \textbf{100.0} & 96.9 & \textbf{100.0} & 96.9 & 96.9 & 96.9 & 62.5 & 75.0 & 96.9 \\
\textsc{aqua} & 94.1 & 78.0 & 76.0 & 76.0 & 75.6 & 76.0 & 73.2 & 74.4 & 79.5 & 79.5 & 78.0 & 75.5 & 77.6 & 79.5 & 63.8 & 63.0 & 68.3 \\
\textsc{asdiv} & 96.7 & 95.8 & 96.0 & 96.0 & 95.9 & 96.0 & 95.9 & 95.5 & \textbf{96.1} & 96.0 & 95.8 & 95.8 & 96.0 & 95.8 & 70.7 & 76.1 & 96.0 \\
\textsc{carp\_en} & 63.1 & 61.8 & 61.7 & 61.9 & \textbf{62.0} & \textbf{62.0} & 61.7 & 61.5 & 61.0 & 61.2 & 61.1 & 61.1 & 61.1 & 61.5 & 36.1 & 39.2 & 61.5 \\
\textsc{cmath} & 95.3 & 92.7 & 93.2 & 93.8 & \textbf{94.0} & \textbf{94.0} & 92.2 & 92.2 & 92.3 & 93.2 & 92.8 & 93.3 & 93.0 & 93.5 & 69.8 & 73.5 & \textbf{94.0} \\
\textsc{cn\_middle\_school} & 82.2 & 78.2 & \textbf{80.2} & 79.2 & 79.2 & 79.2 & 76.2 & 76.2 & \textbf{80.2} & 78.2 & 77.2 & 77.2 & 77.2 & 78.2 & 54.5 & 61.4 & 79.2 \\
\textsc{gaokao\_math\_cloze} & 81.4 & 76.3 & 78.0 & 78.0 & 78.0 & 78.0 & 72.9 & 76.3 & 78.0 & 76.3 & 76.3 & \textbf{79.7} & 78.8 & 77.1 & 47.5 & 57.6 & 75.4 \\
\textsc{gaokao\_math\_qa} & 94.6 & 73.5 & 75.8 & \textbf{77.8} & 77.2 & 77.5 & 56.1 & 73.5 & 68.4 & 70.4 & 72.9 & 72.9 & 70.9 & 63.8 & 52.1 & 56.4 & 75.5 \\
\textsc{gaokao2023en} & 80.8 & 72.2 & 73.0 & 71.9 & 72.2 & 71.4 & 69.9 & 67.0 & 69.4 & 69.6 & 69.6 & 71.7 & 71.7 & 70.1 & 36.6 & 40.5 & \textbf{75.6} \\
\textsc{gaokao2024\_I} & 71.4 & 57.1 & 57.1 & 50.0 & 57.1 & 50.0 & 42.9 & 57.1 & 64.3 & 64.3 & 64.3 & 57.1 & 57.1 & 57.1 & 42.9 & 35.7 & 64.3 \\
\textsc{gaokao2024\_II} & 85.7 & 64.3 & \textbf{71.4} & 64.3 & 57.1 & 57.1 & 50.0 & 50.0 & \textbf{71.4} & 57.1 & 57.1 & 64.3 & 64.3 & 64.3 & 35.7 & 28.6 & 57.1 \\
\textsc{gaokao2024\_mix} & 79.1 & 72.5 & 73.6 & 71.4 & 71.4 & 71.4 & 59.3 & 71.4 & 65.9 & 68.1 & 64.8 & 67.0 & 68.1 & 69.2 & 46.2 & 36.3 & \textbf{79.4} \\
\textsc{gsm8k} & 97.7 & 96.6 & 96.4 & 96.4 & 96.4 & 96.4 & 96.1 & 95.7 & 96.1 & 96.0 & 95.6 & 96.0 & \textbf{96.7} & 96.6 & 46.3 & 54.4 & \textbf{96.7} \\
\textsc{mawps} & 98.8 & \textbf{98.5} & 98.4 & \textbf{98.5} & \textbf{98.5} & \textbf{98.5} & 98.4 & 98.3 & 98.1 & 98.3 & 98.4 & 98.4 & \textbf{98.5} & 98.3 & 72.5 & 81.7 & 98.4 \\
\textsc{minerva\_math} & 48.9 & 41.5 & 39.3 & 40.1 & 40.1 & 39.3 & 37.1 & 36.0 & 38.2 & 38.2 & 39.3 & 37.9 & 40.8 & 40.4 & 14.0 & 15.1 & \textbf{42.1} \\
\textsc{mmlu\_stem} & 90.2 & 72.9 & \textbf{74.0} & 73.4 & 73.7 & 72.9 & 69.0 & 70.2 & 72.0 & 72.4 & 72.2 & \textbf{74.0} & 73.6 & 73.5 & 54.6 & 43.0 & 31.7 \\
\textsc{svamp} & 96.7 & 94.6 & 95.0 & 95.1 & 95.2 & 95.1 & 94.2 & 93.4 & 94.9 & 95.6 & 95.5 & 95.9 & 95.6 & 95.7 & 68.7 & 76.4 & \textbf{96.8} \\
\textsc{tabmwp} & 98.6 & 96.1 & 96.4 & 96.5 & 96.4 & \textbf{96.9} & 95.0 & 94.6 & 95.5 & 96.0 & 95.3 & 96.1 & 96.6 & 96.7 & 63.3 & 68.3 & 96.2 \\
\textsc{olympiadbench} & 61.9 & 44.7 & \textbf{48.4} & 47.3 & 46.1 & 46.7 & 42.1 & 42.7 & 44.1 & 45.5 & 43.3 & 44.9 & 46.1 & 46.1 & 21.6 & 21.9 & 47.9 \\
\textsc{math} & 92.3 & 87.0 & 88.2 & 87.6 & 87.5 & 87.3 & 85.5 & 84.5 & 86.0 & 86.4 & 86.7 & 86.4 & 86.7 & 86.8 & 44.7 & 48.8 & \textbf{89.1} \\
\midrule
\textsc{Average} & 79.8 $\pm$ 4.7 & 71.4 $\pm$ 5.1 & 72.7 $\pm$ 5.1 & 72.1 $\pm$ 5.0 & 72.2 $\pm$ 5.0 & 72.0 $\pm$ 5.0 & 67.6 $\pm$ 5.3 & 68.8 $\pm$ 5.4 & 71.6 $\pm$ 5.0 & 71.2 $\pm$ 5.0 & 70.5 $\pm$ 5.2 & 71.4 $\pm$ 5.1 & 71.8 $\pm$ 4.9 & 71.9 $\pm$ 4.8 & 46.0 $\pm$ 4.1 & 48.1 $\pm$ 4.6 & 71.2 $\pm$ 5.0 \\
\midrule
\textsc{Best Count} & N/A & 1/23 & 5/23 & 2/23 & 4/23 & 5/23 & 0/23 & 0/23 & 4/23 & 0/23 & 1/23 & 3/23 & 1/23 & 1/23 & 0/23 & 0/23 & 6/23 \\
\bottomrule
\end{tabular}
}
\end{table}

\begin{figure}[h]
\centering
\includegraphics[width=0.9\textwidth]{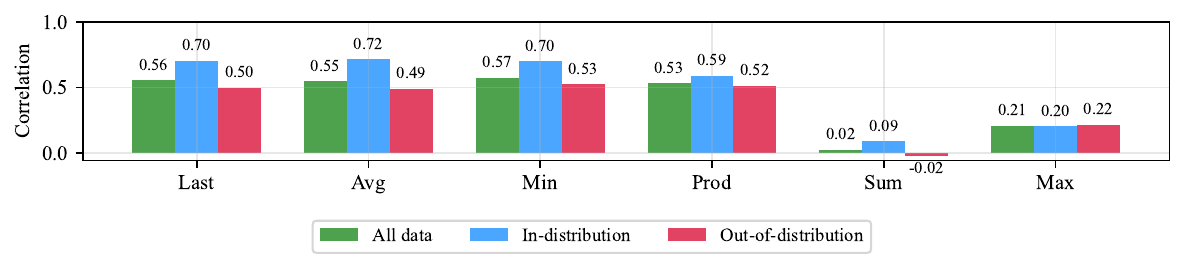}
\caption{
\textbf{Point-biserial correlation between PRM score aggregation methods and solution correctness.} Minimum aggregation shows highest correlation, followed by last reasoning-step scoring. 
Correlation is consistently higher on in-distribution (ID) problems on which the PRM is trained (GSM8K and MATH) than out-of-distribution (OOD) tasks.}
\label{fig:correlation_prm_correctness}
\end{figure}

\begin{figure}[h]
\centering
\includegraphics[width=0.9\textwidth]{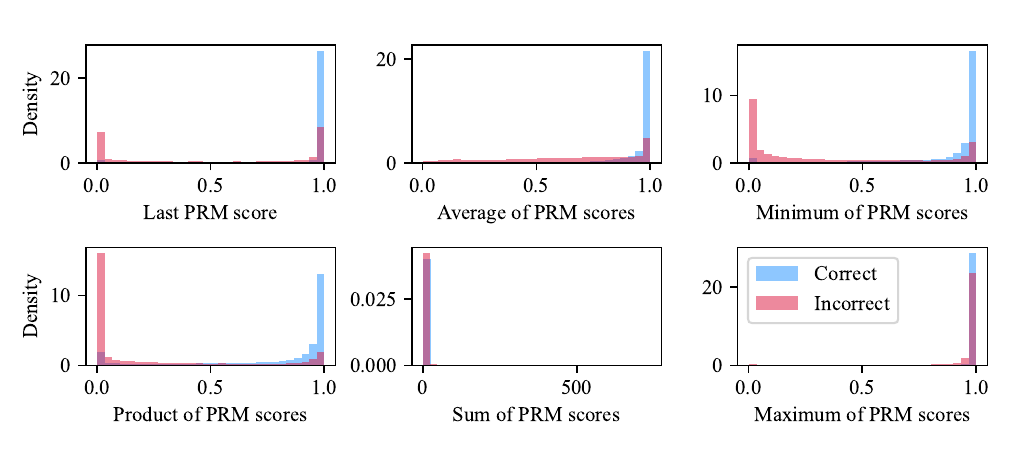}
\caption{
\textbf{PRM score distributions by aggregation method, conditioned on prediction correctness.} Last, average, minimum, and product aggregations effectively separate correct from incorrect predictions.}
\label{fig:all_datasets_distributions}
\end{figure}

\begin{table}[htbp]
\centering
\caption{\textbf{Point-biserial correlations between PRM aggregation methods and prediction correctness by dataset.} Bold indicates best performance per dataset. Correlations vary substantially across problems, with optimal aggregation being dataset-dependent. Overall, minimum aggregation performs best, followed by last-step scoring.}
\medskip  
\label{tab:corr_all_datasets}
\resizebox{0.9\textwidth}{!}{
\begin{tabular}{lcccccc}
\toprule
\textsc{Dataset} & \texttt{BoN\_Last@8} & \texttt{BoN\_Avg@8} & \texttt{BoN\_Min@8} & \texttt{BoN\_Prod@8} & \texttt{BoN\_Sum@8} & \texttt{BoN\_Max@8} \\
\midrule
\textsc{aime25} & 0.494 & 0.474 & \textbf{0.582} & 0.444 & 0.283 & 0.095 \\
\textsc{aime24} & 0.623 & 0.597 & \textbf{0.680} & 0.487 & 0.529 & 0.151 \\
\textsc{amc23} & 0.668 & 0.700 & \textbf{0.726} & 0.626 & 0.170 & 0.280 \\
\textsc{sat\_math} & \textbf{0.690} & 0.375 & 0.487 & 0.330 & 0.069 & 0.039 \\
\textsc{aqua} & 0.309 & 0.309 & \textbf{0.342} & 0.306 & 0.106 & 0.291 \\
\textsc{asdiv} & 0.447 & \textbf{0.466} & 0.450 & 0.424 & -0.049 & 0.116 \\
\textsc{carp\_en} & 0.137 & 0.149 & \textbf{0.154} & 0.150 & -0.026 & 0.060 \\
\textsc{cmath} & 0.557 & 0.559 & 0.627 & \textbf{0.637} & 0.063 & 0.200 \\
\textsc{cn\_middle\_school} & 0.432 & 0.419 & \textbf{0.465} & 0.444 & -0.161 & 0.119 \\
\textsc{gaokao\_math\_cloze} & 0.455 & 0.500 & \textbf{0.555} & 0.482 & 0.094 & 0.098 \\
\textsc{gaokao\_math\_qa} & 0.510 & 0.454 & \textbf{0.546} & 0.488 & -0.016 & 0.216 \\
\textsc{gaokao2023en} & 0.561 & 0.590 & \textbf{0.602} & 0.540 & 0.148 & 0.175 \\
\textsc{gaokao2024\_I} & 0.285 & 0.275 & \textbf{0.438} & 0.337 & -0.064 & 0.126 \\
\textsc{gaokao2024\_II} & 0.376 & 0.447 & 0.545 & \textbf{0.584} & 0.048 & 0.094 \\
\textsc{gaokao2024\_mix} & 0.482 & 0.422 & \textbf{0.495} & 0.343 & 0.081 & 0.187 \\
\textsc{gsm8k} & 0.560 & 0.595 & \textbf{0.597} & 0.565 & 0.099 & 0.120 \\
\textsc{mawps} & 0.333 & \textbf{0.379} & 0.322 & 0.296 & 0.015 & 0.091 \\
\textsc{minerva\_math} & 0.349 & 0.416 & \textbf{0.477} & 0.474 & -0.022 & 0.183 \\
\textsc{mmlu\_stem} & \textbf{0.360} & 0.280 & 0.281 & 0.266 & 0.085 & 0.182 \\
\textsc{svamp} & 0.700 & \textbf{0.706} & 0.694 & 0.673 & 0.109 & 0.239 \\
\textsc{tabmwp} & 0.542 & \textbf{0.591} & 0.588 & 0.570 & 0.090 & 0.211 \\
\textsc{olympiadbench} & 0.571 & 0.603 & \textbf{0.635} & 0.578 & 0.213 & 0.145 \\
\textsc{math} & 0.708 & \textbf{0.718} & 0.703 & 0.585 & 0.130 & 0.200 \\
\midrule
\textsc{Best Count} & 2/23 & 5/23 & 14/23 & 2/23 & 0/23 & 0/23 \\
\midrule 
\textsc{Average} & 0.557 & 0.550 & 0.575 & 0.533 & 0.024 & 0.209 \\
\bottomrule
\end{tabular}
}
\end{table}

%% file: neurips_2025.bbl
\begin{thebibliography}{27}
\providecommand{\natexlab}[1]{#1}
\providecommand{\url}[1]{\texttt{#1}}
\expandafter\ifx\csname urlstyle\endcsname\relax
  \providecommand{\doi}[1]{doi: #1}\else
  \providecommand{\doi}{doi: \begingroup \urlstyle{rm}\Url}\fi

\bibitem[Wei et~al.(2022)Wei, Wang, Schuurmans, Bosma, Ichter, Xia, Chi, Le, and Zhou]{wei2022cot}
Jason Wei, Xuezhi Wang, Dale Schuurmans, Maarten Bosma, Brian Ichter, Fei Xia, Ed~H. Chi, Quoc~V. Le, and Denny Zhou.
\newblock Chain-of-thought prompting elicits reasoning in large language models.
\newblock In \emph{Proceedings of the 36th International Conference on Neural Information Processing Systems}, NIPS '22, Red Hook, NY, USA, 2022. Curran Associates Inc.
\newblock ISBN 9781713871088.

\bibitem[Sprague et~al.(2025)Sprague, Yin, Rodriguez, Jiang, Wadhwa, Singhal, Zhao, Ye, Mahowald, and Durrett]{sprague2025to}
Zayne~Rea Sprague, Fangcong Yin, Juan~Diego Rodriguez, Dongwei Jiang, Manya Wadhwa, Prasann Singhal, Xinyu Zhao, Xi~Ye, Kyle Mahowald, and Greg Durrett.
\newblock To cot or not to cot? chain-of-thought helps mainly on math and symbolic reasoning.
\newblock In \emph{The Thirteenth International Conference on Learning Representations}, 2025.
\newblock URL \url{https://openreview.net/forum?id=w6nlcS8Kkn}.

\bibitem[Zhang et~al.(2025)Zhang, Zheng, Wu, Zhang, Lin, Yu, Liu, Zhou, and Lin]{zhang2025lessonsdevelopingprocessreward}
Zhenru Zhang, Chujie Zheng, Yangzhen Wu, Beichen Zhang, Runji Lin, Bowen Yu, Dayiheng Liu, Jingren Zhou, and Junyang Lin.
\newblock The lessons of developing process reward models in mathematical reasoning, 2025.
\newblock URL \url{https://arxiv.org/abs/2501.07301}.

\bibitem[Yang et~al.(2024)Yang, Zhang, Hui, Gao, Yu, Li, Liu, Tu, Zhou, Lin, Lu, Xue, Lin, Liu, Ren, and Zhang]{yang2024qwen25mathtechnicalreportmathematical}
An~Yang, Beichen Zhang, Binyuan Hui, Bofei Gao, Bowen Yu, Chengpeng Li, Dayiheng Liu, Jianhong Tu, Jingren Zhou, Junyang Lin, Keming Lu, Mingfeng Xue, Runji Lin, Tianyu Liu, Xingzhang Ren, and Zhenru Zhang.
\newblock Qwen2.5-math technical report: Toward mathematical expert model via self-improvement, 2024.
\newblock URL \url{https://arxiv.org/abs/2409.12122}.

\bibitem[Lightman et~al.(2024)Lightman, Kosaraju, Burda, Edwards, Baker, Lee, Leike, Schulman, Sutskever, and Cobbe]{lightman2024lets}
Hunter Lightman, Vineet Kosaraju, Yuri Burda, Harrison Edwards, Bowen Baker, Teddy Lee, Jan Leike, John Schulman, Ilya Sutskever, and Karl Cobbe.
\newblock Let's verify step by step.
\newblock In \emph{The Twelfth International Conference on Learning Representations}, 2024.
\newblock URL \url{https://openreview.net/forum?id=v8L0pN6EOi}.

\bibitem[Uesato et~al.(2022)Uesato, Kushman, Kumar, Song, Siegel, Wang, Creswell, Irving, and Higgins]{uesato2022solvingmathwordproblems}
Jonathan Uesato, Nate Kushman, Ramana Kumar, Francis Song, Noah Siegel, Lisa Wang, Antonia Creswell, Geoffrey Irving, and Irina Higgins.
\newblock Solving math word problems with process- and outcome-based feedback, 2022.
\newblock URL \url{https://arxiv.org/abs/2211.14275}.

\bibitem[Muennighoff et~al.(2025)Muennighoff, Yang, Shi, Li, Fei-Fei, Hajishirzi, Zettlemoyer, Liang, Candes, and Hashimoto]{muennighoff2025s1}
Niklas Muennighoff, Zitong Yang, Weijia Shi, Xiang~Lisa Li, Li~Fei-Fei, Hannaneh Hajishirzi, Luke Zettlemoyer, Percy Liang, Emmanuel Candes, and Tatsunori Hashimoto.
\newblock s1: Simple test-time scaling.
\newblock In \emph{Workshop on Reasoning and Planning for Large Language Models}, 2025.
\newblock URL \url{https://openreview.net/forum?id=LdH0vrgAHm}.

\bibitem[DeepSeek-AI et~al.(2025)DeepSeek-AI, Guo, Yang, Zhang, Song, Zhang, Xu, Zhu, Ma, Wang, Bi, Zhang, Yu, Wu, Wu, Gou, Shao, Li, Gao, Liu, Xue, Wang, Wu, Feng, Lu, Zhao, Deng, Zhang, Ruan, Dai, Chen, Ji, Li, Lin, Dai, Luo, Hao, Chen, Li, Zhang, Bao, Xu, Wang, Ding, Xin, Gao, Qu, Li, Guo, Li, Wang, Chen, Yuan, Qiu, Li, Cai, Ni, Liang, Chen, Dong, Hu, Gao, Guan, Huang, Yu, Wang, Zhang, Zhao, Wang, Zhang, Xu, Xia, Zhang, Zhang, Tang, Li, Wang, Li, Tian, Huang, Zhang, Wang, Chen, Du, Ge, Zhang, Pan, Wang, Chen, Jin, Chen, Lu, Zhou, Chen, Ye, Wang, Yu, Zhou, Pan, Li, Zhou, Wu, Ye, Yun, Pei, Sun, Wang, Zeng, Zhao, Liu, Liang, Gao, Yu, Zhang, Xiao, An, Liu, Wang, Chen, Nie, Cheng, Liu, Xie, Liu, Yang, Li, Su, Lin, Li, Jin, Shen, Chen, Sun, Wang, Song, Zhou, Wang, Shan, Li, Wang, Wei, Zhang, Xu, Li, Zhao, Sun, Wang, Yu, Zhang, Shi, Xiong, He, Piao, Wang, Tan, Ma, Liu, Guo, Ou, Wang, Gong, Zou, He, Xiong, Luo, You, Liu, Zhou, Zhu, Xu, Huang, Li, Zheng, Zhu, Ma, Tang, Zha, Yan, Ren, Ren, Sha, Fu, Xu, Xie, Zhang, Hao, Ma, Yan, Wu, Gu, Zhu, Liu, Li, Xie, Song, Pan, Huang, Xu, Zhang, and Zhang]{deepseekai2025deepseekr1incentivizingreasoningcapability}
DeepSeek-AI, Daya Guo, Dejian Yang, Haowei Zhang, Junxiao Song, Ruoyu Zhang, Runxin Xu, Qihao Zhu, Shirong Ma, Peiyi Wang, Xiao Bi, Xiaokang Zhang, Xingkai Yu, Yu~Wu, Z.~F. Wu, Zhibin Gou, Zhihong Shao, Zhuoshu Li, Ziyi Gao, Aixin Liu, Bing Xue, Bingxuan Wang, Bochao Wu, Bei Feng, Chengda Lu, Chenggang Zhao, Chengqi Deng, Chenyu Zhang, Chong Ruan, Damai Dai, Deli Chen, Dongjie Ji, Erhang Li, Fangyun Lin, Fucong Dai, Fuli Luo, Guangbo Hao, Guanting Chen, Guowei Li, H.~Zhang, Han Bao, Hanwei Xu, Haocheng Wang, Honghui Ding, Huajian Xin, Huazuo Gao, Hui Qu, Hui Li, Jianzhong Guo, Jiashi Li, Jiawei Wang, Jingchang Chen, Jingyang Yuan, Junjie Qiu, Junlong Li, J.~L. Cai, Jiaqi Ni, Jian Liang, Jin Chen, Kai Dong, Kai Hu, Kaige Gao, Kang Guan, Kexin Huang, Kuai Yu, Lean Wang, Lecong Zhang, Liang Zhao, Litong Wang, Liyue Zhang, Lei Xu, Leyi Xia, Mingchuan Zhang, Minghua Zhang, Minghui Tang, Meng Li, Miaojun Wang, Mingming Li, Ning Tian, Panpan Huang, Peng Zhang, Qiancheng Wang, Qinyu Chen, Qiushi Du, Ruiqi Ge, Ruisong Zhang, Ruizhe Pan, Runji Wang, R.~J. Chen, R.~L. Jin, Ruyi Chen, Shanghao Lu, Shangyan Zhou, Shanhuang Chen, Shengfeng Ye, Shiyu Wang, Shuiping Yu, Shunfeng Zhou, Shuting Pan, S.~S. Li, Shuang Zhou, Shaoqing Wu, Shengfeng Ye, Tao Yun, Tian Pei, Tianyu Sun, T.~Wang, Wangding Zeng, Wanjia Zhao, Wen Liu, Wenfeng Liang, Wenjun Gao, Wenqin Yu, Wentao Zhang, W.~L. Xiao, Wei An, Xiaodong Liu, Xiaohan Wang, Xiaokang Chen, Xiaotao Nie, Xin Cheng, Xin Liu, Xin Xie, Xingchao Liu, Xinyu Yang, Xinyuan Li, Xuecheng Su, Xuheng Lin, X.~Q. Li, Xiangyue Jin, Xiaojin Shen, Xiaosha Chen, Xiaowen Sun, Xiaoxiang Wang, Xinnan Song, Xinyi Zhou, Xianzu Wang, Xinxia Shan, Y.~K. Li, Y.~Q. Wang, Y.~X. Wei, Yang Zhang, Yanhong Xu, Yao Li, Yao Zhao, Yaofeng Sun, Yaohui Wang, Yi~Yu, Yichao Zhang, Yifan Shi, Yiliang Xiong, Ying He, Yishi Piao, Yisong Wang, Yixuan Tan, Yiyang Ma, Yiyuan Liu, Yongqiang Guo, Yuan Ou, Yuduan Wang, Yue Gong, Yuheng Zou, Yujia He, Yunfan Xiong, Yuxiang Luo, Yuxiang You, Yuxuan Liu, Yuyang Zhou, Y.~X. Zhu, Yanhong Xu, Yanping Huang, Yaohui Li, Yi~Zheng, Yuchen Zhu, Yunxian Ma, Ying Tang, Yukun Zha, Yuting Yan, Z.~Z. Ren, Zehui Ren, Zhangli Sha, Zhe Fu, Zhean Xu, Zhenda Xie, Zhengyan Zhang, Zhewen Hao, Zhicheng Ma, Zhigang Yan, Zhiyu Wu, Zihui Gu, Zijia Zhu, Zijun Liu, Zilin Li, Ziwei Xie, Ziyang Song, Zizheng Pan, Zhen Huang, Zhipeng Xu, Zhongyu Zhang, and Zhen Zhang.
\newblock Deepseek-r1: Incentivizing reasoning capability in llms via reinforcement learning, 2025.
\newblock URL \url{https://arxiv.org/abs/2501.12948}.

\bibitem[Yao et~al.(2023)Yao, Yu, Zhao, Shafran, Griffiths, Cao, and Narasimhan]{yao2023tree}
Shunyu Yao, Dian Yu, Jeffrey Zhao, Izhak Shafran, Thomas~L. Griffiths, Yuan Cao, and Karthik Narasimhan.
\newblock {Tree of Thoughts}: Deliberate problem solving with large language models, 2023.

\bibitem[Weitzman(1979)]{weitzman1979pandora}
Martin~L. Weitzman.
\newblock Optimal search for the best alternative.
\newblock \emph{Econometrica}, 47\penalty0 (3):\penalty0 641--654, 1979.
\newblock URL \url{https://onlinelibrary.wiley.com/doi/abs/0012-9682(197905)47:3&lt;641:OSFTBA&gt;2.0.CO;2-1}.

\bibitem[Xie et~al.(2024{\natexlab{a}})Xie, Astudillo, Frazier, Scully, and Terenin]{xie2024costaware}
Qian Xie, Raul Astudillo, Peter~I. Frazier, Ziv Scully, and Alexander Terenin.
\newblock Cost-aware bayesian optimization via the pandora's box gittins index.
\newblock In \emph{The Thirty-eighth Annual Conference on Neural Information Processing Systems}, 2024{\natexlab{a}}.
\newblock URL \url{https://openreview.net/forum?id=Ouc1F0Sfb7}.

\bibitem[Jones et~al.(1998)Jones, Schonlau, and Welch]{jones1998EI}
Donald Jones, Matthias Schonlau, and William Welch.
\newblock Efficient global optimization of expensive black-box functions.
\newblock \emph{Journal of Global Optimization}, 13:\penalty0 455--492, 12 1998.
\newblock \doi{10.1023/A:1008306431147}.

\bibitem[Garnett(2023)]{garnett_bayesoptbook_2023}
Roman Garnett.
\newblock \emph{{Bayesian Optimization}}.
\newblock Cambridge University Press, 2023.

\bibitem[Balunović et~al.(2025)Balunović, Dekoninck, Petrov, Jovanović, and Vechev]{balunovic_srimatharena_2025}
Mislav Balunović, Jasper Dekoninck, Ivo Petrov, Nikola Jovanović, and Martin Vechev.
\newblock Matharena: Evaluating llms on uncontaminated math competitions, February 2025.
\newblock URL \url{https://matharena.ai/}.

\bibitem[Hochlehnert et~al.(2025)Hochlehnert, Bhatnagar, Udandarao, Albanie, Prabhu, and Bethge]{hochlehnert2025soberlookprogresslanguage}
Andreas Hochlehnert, Hardik Bhatnagar, Vishaal Udandarao, Samuel Albanie, Ameya Prabhu, and Matthias Bethge.
\newblock A sober look at progress in language model reasoning: Pitfalls and paths to reproducibility, 2025.
\newblock URL \url{https://arxiv.org/abs/2504.07086}.

\bibitem[Zhang et~al.(2024{\natexlab{a}})Zhang, Huang, Zhou, Li, and Ouyang]{zhang2024accessinggpt4levelmathematical}
Di~Zhang, Xiaoshui Huang, Dongzhan Zhou, Yuqiang Li, and Wanli Ouyang.
\newblock Accessing gpt-4 level mathematical olympiad solutions via monte carlo tree self-refine with llama-3 8b, 2024{\natexlab{a}}.
\newblock URL \url{https://arxiv.org/abs/2406.07394}.

\bibitem[Hao et~al.(2023)Hao, Gu, Ma, Hong, Wang, Wang, and Hu]{hao2023reasoning}
Shibo Hao, Yi~Gu, Haodi Ma, Joshua~Jiahua Hong, Zhen Wang, Daisy~Zhe Wang, and Zhiting Hu.
\newblock Reasoning with language model is planning with world model.
\newblock In \emph{The 2023 Conference on Empirical Methods in Natural Language Processing}, 2023.
\newblock URL \url{https://openreview.net/forum?id=VTWWvYtF1R}.

\bibitem[Xie et~al.(2023)Xie, Kawaguchi, Zhao, Zhao, Kan, He, and Xie]{xie2023selfevaluation}
Yuxi Xie, Kenji Kawaguchi, Yiran Zhao, Xu~Zhao, Min-Yen Kan, Junxian He, and Qizhe Xie.
\newblock Self-evaluation guided beam search for reasoning.
\newblock In \emph{Thirty-seventh Conference on Neural Information Processing Systems}, 2023.
\newblock URL \url{https://openreview.net/forum?id=Bw82hwg5Q3}.

\bibitem[Zhang et~al.(2024{\natexlab{b}})Zhang, Zhoubian, Hu, Yue, Dong, and Tang]{zhang2024restmcts}
Dan Zhang, Sining Zhoubian, Ziniu Hu, Yisong Yue, Yuxiao Dong, and Jie Tang.
\newblock Re{ST}-{MCTS}*: {LLM} self-training via process reward guided tree search.
\newblock In \emph{The Thirty-eighth Annual Conference on Neural Information Processing Systems}, 2024{\natexlab{b}}.
\newblock URL \url{https://openreview.net/forum?id=8rcFOqEud5}.

\bibitem[Wan et~al.(2024)Wan, Feng, Wen, McAleer, Wen, Zhang, and Wang]{wan2024alphats}
Ziyu Wan, Xidong Feng, Muning Wen, Stephen~Marcus McAleer, Ying Wen, Weinan Zhang, and Jun Wang.
\newblock Alphazero-like tree-search can guide large language model decoding and training.
\newblock In \emph{Proceedings of the 41st International Conference on Machine Learning}, ICML'24. JMLR.org, 2024.

\bibitem[Xie et~al.(2024{\natexlab{b}})Xie, Goyal, Zheng, Kan, Lillicrap, Kawaguchi, and Shieh]{xie2024mctsllm}
Yuxi Xie, Anirudh Goyal, Wenyue Zheng, Min-Yen Kan, Timothy~P Lillicrap, Kenji Kawaguchi, and Michael Shieh.
\newblock Monte carlo tree search boosts reasoning via iterative preference learning.
\newblock \emph{arXiv preprint arXiv:2405.00451}, 2024{\natexlab{b}}.

\bibitem[Guan et~al.(2025)Guan, Zhang, Liu, Shang, Sun, Zhu, Yang, and Yang]{guan2025rstarmath}
Xinyu Guan, Li~Lyna Zhang, Yifei Liu, Ning Shang, Youran Sun, Yi~Zhu, Fan Yang, and Mao Yang.
\newblock rstar-math: Small {LLM}s can master math reasoning with self-evolved deep thinking.
\newblock In \emph{Forty-second International Conference on Machine Learning}, 2025.
\newblock URL \url{https://openreview.net/forum?id=5zwF1GizFa}.

\bibitem[Team(2024)]{qwen25}
Qwen Team.
\newblock Qwen2.5 technical report.
\newblock \emph{arXiv preprint arXiv:2412.15115}, 2024.

\bibitem[Virtanen et~al.(2020)Virtanen, Gommers, Oliphant, Haberland, Reddy, Cournapeau, Burovski, Peterson, Weckesser, Bright, {van der Walt}, Brett, Wilson, Millman, Mayorov, Nelson, Jones, Kern, Larson, Carey, Polat, Feng, Moore, {VanderPlas}, Laxalde, Perktold, Cimrman, Henriksen, Quintero, Harris, Archibald, Ribeiro, Pedregosa, {van Mulbregt}, and {SciPy 1.0 Contributors}]{pauli2020scipy}
Pauli Virtanen, Ralf Gommers, Travis~E. Oliphant, Matt Haberland, Tyler Reddy, David Cournapeau, Evgeni Burovski, Pearu Peterson, Warren Weckesser, Jonathan Bright, St{\'e}fan~J. {van der Walt}, Matthew Brett, Joshua Wilson, K.~Jarrod Millman, Nikolay Mayorov, Andrew R.~J. Nelson, Eric Jones, Robert Kern, Eric Larson, C~J Carey, {\.I}lhan Polat, Yu~Feng, Eric~W. Moore, Jake {VanderPlas}, Denis Laxalde, Josef Perktold, Robert Cimrman, Ian Henriksen, E.~A. Quintero, Charles~R. Harris, Anne~M. Archibald, Ant{\^o}nio~H. Ribeiro, Fabian Pedregosa, Paul {van Mulbregt}, and {SciPy 1.0 Contributors}.
\newblock {{SciPy} 1.0: Fundamental Algorithms for Scientific Computing in Python}.
\newblock \emph{Nature Methods}, 17:\penalty0 261--272, 2020.
\newblock \doi{10.1038/s41592-019-0686-2}.

\bibitem[Hagberg et~al.(2008)Hagberg, Schult, and Swart]{networkx}
Aric~A. Hagberg, Daniel~A. Schult, and Pieter~J. Swart.
\newblock Exploring network structure, dynamics, and function using networkx.
\newblock In Ga\"el Varoquaux, Travis Vaught, and Jarrod Millman, editors, \emph{Proceedings of the 7th Python in Science Conference}, pages 11 -- 15, Pasadena, CA USA, 2008.

\bibitem[Kwon et~al.(2023)Kwon, Li, Zhuang, Sheng, Zheng, Yu, Gonzalez, Zhang, and Stoica]{kwon2023efficient}
Woosuk Kwon, Zhuohan Li, Siyuan Zhuang, Ying Sheng, Lianmin Zheng, Cody~Hao Yu, Joseph~E. Gonzalez, Hao Zhang, and Ion Stoica.
\newblock Efficient memory management for large language model serving with pagedattention.
\newblock In \emph{Proceedings of the ACM SIGOPS 29th Symposium on Operating Systems Principles}, 2023.

\bibitem[{Vector Institute}(2025)]{vectorinference2025}
{Vector Institute}.
\newblock {Vector-Inference}: Efficient llm inference on slurm clusters, 2025.
\newblock URL \url{https://github.com/VectorInstitute/vector-inference}.
\newblock GitHub repository, accessed 2025-09-04.

\end{thebibliography}
